\documentclass[runningheads]{llncs}
\usepackage[T1]{fontenc}
\usepackage{graphicx,verbatim}
\usepackage{amsmath}
\usepackage{amssymb}
\usepackage{hyperref}
\usepackage{xcolor}
\usepackage{tabularx}
\usepackage{multirow}
\usepackage{pifont}
\usepackage{indentfirst}
\usepackage{amssymb}
\usepackage{marvosym}

\begin{document}
\title{Detecting Clinical Hallucinations in LVLMs via Counterfactual Visual Grounding Uncertainty}
\titlerunning{Detecting Clinical Hallucinations via CF Visual Grounding Uncertainty} 

\author{Xiao Song\inst{1} \and 
Haonan Qin\inst{1} \and 
Zhaoxu Zhang \inst{1} \and 
Jiong Zhang\inst{3} \and 
Yuqi Fang \inst{1,2}\textsuperscript{(\Letter)} \and 
Caifeng Shan \inst{1}\textsuperscript{(\Letter)} 
}
\authorrunning{X. Song et al.}
\institute{School of Intelligent Science and Technology, Nanjing University, Suzhou, China \\
\email{\{yqfang, cfshan\}@nju.edu.cn} \and
National Institute of Healthcare Data Science at Nanjing University, Nanjing University, Nanjing, China \and
School of Biomedical Engineering, Nanjing University, Suzhou, China
}
    
\maketitle  
{
    \renewcommand{\thefootnote}{} 
    \footnotetext{\noindent X. Song and H. Qin---Contributed Equally.}
}

\begin{abstract}
Large vision-language models (LVLMs) are increasingly used for clinical image understanding, yet they remain vulnerable to \emph{hallucinations}--producing textual findings or attributes not supported by the image. We present a vision-traceable hallucination detection framework that audits arbitrary LVLM responses via visual evidence grounding, requiring neither modification nor internal access to the hidden states of LVLMs. Given an LVLM response, we extract visually verifiable entities and use a medical-domain-adapted Qwen-VL grounding verifier to localize each entity on the input image. To enhance the robustness of our detection method, we introduce a counterfactual entity perturbation method and estimate visual evidence uncertainty by contrasting factual and counterfactual grounding results. Specifically, we compute an entity-level uncertainty score from the positive confidence, counterfactual confidence, and their grounding overlap for binary hallucination decision-making. Experiments on multiple medical imaging modalities and LVLM backbones demonstrate that our method consistently improves hallucination detection performance over recent baselines, while providing interpretable localization evidence and strong cross-model transferability. Code and dataset are available at \url{https://github.com/Agentic-CliniAI/CounterVHD}.

\end{abstract}

\begin{figure}[t]
  \begin{center}
    \centerline{\includegraphics[width=\columnwidth]{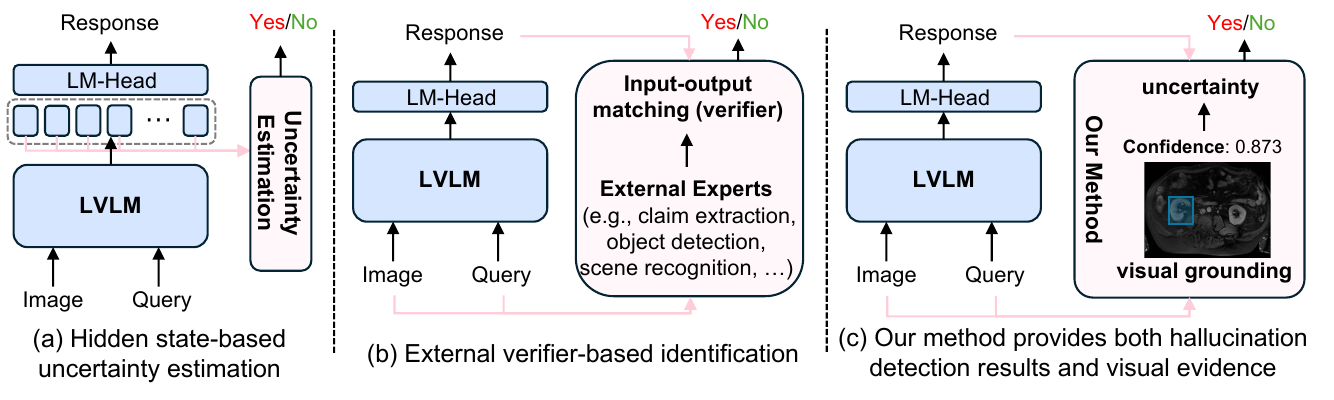}}
    \caption{
Comparison of hallucination detection paradigms. 
(a) \textbf{Hidden state-based methods}: access LVLM's internal hidden states. (b) \textbf{External verifier-based methods}: rely on external expert models. (c) \textbf{Ours}: identifies hallucinations by aligning responses to visual regions, improving interpretability via visual evidence.
    }
    \label{fig:fig0}
  \end{center}
\end{figure}

\section{Introduction}
Medical large vision--language models (LVLMs) have demonstrated impressive capability in interpreting images and generating free-form responses for a wide range of clinical tasks, such as visual question answering, report generation, and decision support.
Despite rapid progress, LVLMs remain prone to \emph{hallucinations}~\cite{khanal2025hallucination,liao2025vision,chen2024detecting,xiao2025detecting,song2024rethinking}: they may describe findings, anatomy, or devices that are not supported by the input image or assign incorrect attributes (e.g., laterality, location, severity). 
In safety-critical medical scenarios, such errors can mislead clinicians and undermine trust, motivating reliable auditing tools.

Recent research on hallucination detection in LVLMs can be grouped into two paradigms~\cite{liao2025vision}, as depicted in Fig.~\ref{fig:fig0} (a) and (b).
(a) \textbf{Hidden state-based methods}~\cite{hardy2024rextrust,li2024reference,zou2025uncertainty} apply LVLM's internal hidden state  for predictive uncertainty as a proxy for hallucination rate, where higher uncertainty signals higher likelihood of erroneous generation. While effective, their applicability is often constrained by the black-box nature of most commercial models, which do not grant users access to their internal states.
(b) \textbf{External verifier-based methods}~\cite{chen2024detecting,jing2024faithscore,wang2023amber,liao2025univrse} deploy well-performed LVLMs or specifically trained expert models to verify whether the outputs align faithfully with both the query and visual evidence.
However, beyond binary detection conclusions, explicitly localizing the corresponding visual evidence is essential for validating findings and ensuring diagnostic tracing, which is a critical capability that existing methods neglected. 

\begin{figure}[t]
  \begin{center}
    \centerline{\includegraphics[width=.9\columnwidth]{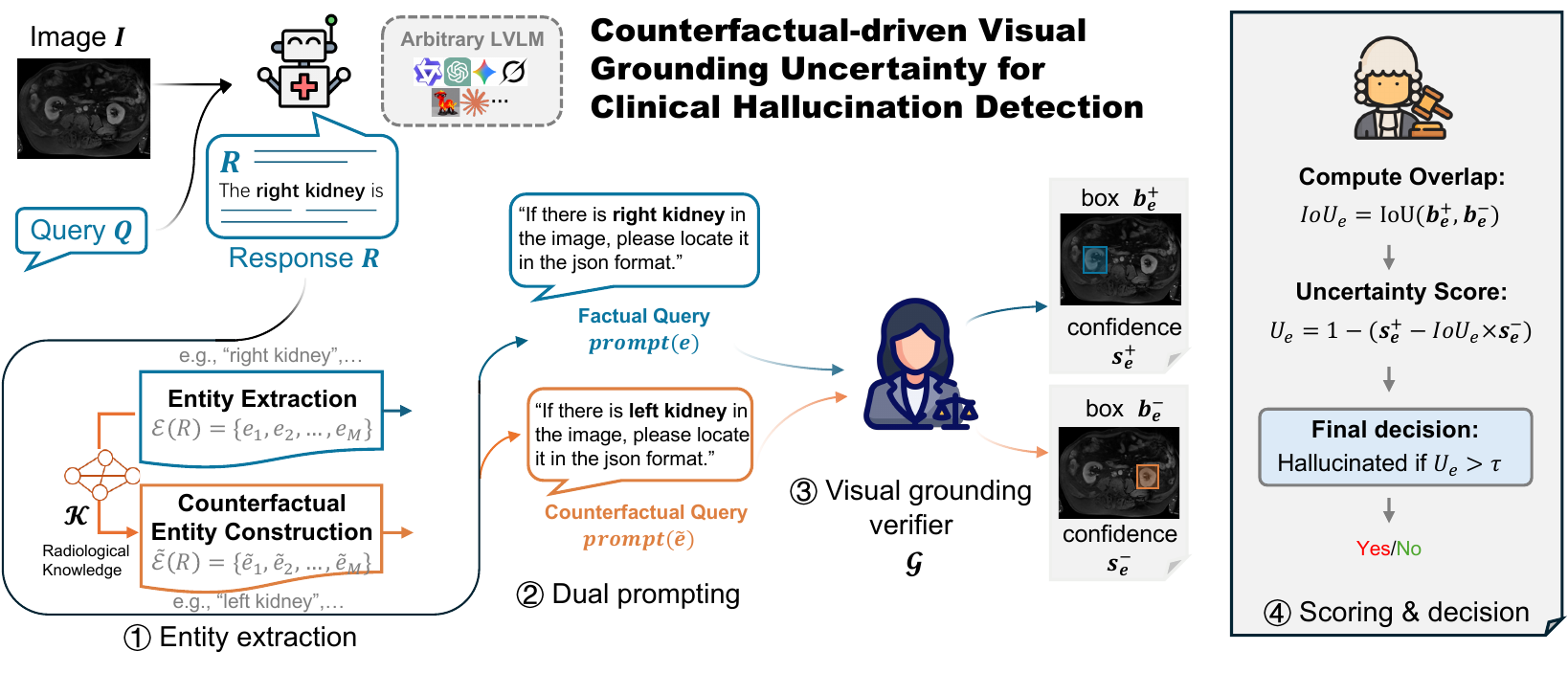}}
    \caption{
Pipeline of the proposed Counterfactual-driven Visual Grounding Uncertainty Estimation method. 
\ding{172} Given a response $R$ from an arbitrary LVLM, we extract entities $\mathcal{E}$ and construct counterfactual entities $\tilde{\mathcal{E}}$ using radiological knowledge.
\ding{173} Constructing the factual and counterfactual queries.
\ding{174} A trained grounding verifier predicts bounding boxes ($b_e^+, b_e^-$) and confidence scores ($s_e^+, s_e^-$).
\ding{175} Uncertainty estimation and the final decision is made by comparing the aggregated score against threshold.
    }
    \label{fig:main}
  \end{center}
\end{figure}

To tackle this problem, in this paper, we propose a vision-aware hallucination detection method via Counterfactual-driven Visual Grounding Uncertainty Estimation (see Fig.~\ref{fig:fig0} (c) and Fig.~\ref{fig:main}).
Specifically, we extract entity mentions from responses and localize them using a Qwen-VL \emph{visual grounding} verifier fine-tuned specifically for the medical domain.
Furthermore, to enhance robustness of our method, we introduce a \emph{counterfactual entity perturbation branch} that prompts the verifier to ground a hard negative counterpart that is semantically related yet spatially non-overlapping from the original factual entity.
Subsequently, we derive a vision-aware uncertainty score by comparing the IoU-reweighted grounding confidences of factual and counterfactual bounding boxes, thereby enabling robust and accurate hallucination detection.
Extensive experiments across diverse medical imaging modalities and LVLM backbones demonstrate that our method achieves superior performance with robust generalizability over recent baselines, while our vision-aware design simultaneously provides interpretable visual evidence for clinical auditing. Our main contributions are threefold:
\begin{itemize}
    \item We introduce a vision-aware hallucination detection method for LVLMs in clinical scenarios. By leveraging visual grounding, the method provides clinicians with both hallucination decisions and precise entity-level localization.
    \item We propose a counterfactual-driven uncertainty estimation method that identifies spurious evidence by contrasting factual and counterfactual grounding, ensuring robust hallucination identification.
    \item Experiments across diverse medical modalities and LVLM backbones demonstrate that our method achieves superior performance while ensuring robust generalizability and visual interpretability.
\end{itemize}

\section{Methodology}
\subsection{Problem Setup and Overview}
In this paper, we study plug-and-play detection of hallucinations in the responses of arbitrary LVLMs in clinical scenarios.
Given an image $I$ and a black-box LVLM $\mathcal{M}$, the model generates a response $R=\mathcal{M}(I,Q)$ to a user query $Q$.
Our goal is to decide whether $R$ contains \emph{visual-grounded claims} that contradict the visual facts in $I$.
To this end, we introduce an external visual grounding verifier $\mathcal{G}$ (instantiated by a fine-tuned Qwen3-VL-2B~\cite{Qwen3-VL} model),
which is decoupled from $\mathcal{M}$ and thus generalizes across arbitrary LVLMs.

The key idea is to convert hallucination detection into \emph{visual evidence alignment}:
As illustrated in Fig.~\ref{fig:main}, we extract entity mentions from $R$, ground them onto $I$ to obtain factual evidence, and additionally construct \emph{counterfactual} entity descriptions that should not visually overlap with the original entity.
We then estimate a test-time uncertainty by contrasting the factual and counterfactual grounding results, yielding a hallucination probability.

\subsection{Entity Extraction from LVLM Responses}
We first extract a set of entity mentions from $R$ corresponding to visually verifiable concepts in medical images (e.g., anatomy, lesions):
\begin{equation}
    \mathcal{E}(R)=\{e_\mathrm{m}\}_{\mathrm{m}=1}^{\mathrm{M}},
\end{equation}
where $\mathrm{M}$ is the number of extracted entities.
In practice, we prompt the GPT-4.1-mini model via API as a sophisticated text parser.

\subsection{Visual Grounding Verifier}
For each entity $e\in\mathcal{E}(R)$, we prompt the grounding model $\mathcal{G}$ to return a bounding box and a confidence score:
\begin{equation}
    (b^{+}_e, s^{+}_e)=\mathcal{G}(I, \text{prompt}(e)),
\end{equation}
where $b^{+}_e\in\mathbb{R}^{4}$ denotes the predicted box and $s^{+}_e\in[0,1]$ indicates grounding confidence.
Specifically, the grounding prediction $b^{+}_e$ is obtained from a single deterministic inference of $\mathcal{G}$.
To compute $s^{+}_e$, we consider two alternative algorithms: (a) \textbf{Monte-Carlo Sampling}: Under elevated temperature and stochastic sampling, $\mathcal{G}$ generates $\mathrm{N}$ diverse candidate boxes $\{b^{+(\mathrm{n})}_e\}_{\mathrm{n}=1}^\mathrm{N}$; $s^{+}_e$ is defined as their pairwise $\mathrm{mIoU}$:
\begin{equation}
    s^{+}_e=\frac{2}{\mathrm{N}(\mathrm{N}-1)}\sum_{1\leq i<j\leq\mathrm{N}}\mathrm{IoU}(b^{+(i)}_e,b^{+(j)}_e).
\end{equation}
(b) \textbf{Internal Logit Probability}: Compute $s^{+}_e$ by averaging the softmax probabilities of the predicted box logits via the hidden state of $\mathcal{G}$. 
Note that different from hidden state-based methods that require internal LVLM access, our method leverages the internal states of $\mathcal{G}$ while remaining model-agnostic, thereby ensuring generalizability across various LVLMs.

\subsection{Counterfactual Entity Construction}
\label{sec:counterfactual}
Grounding confidence alone is insufficient. Obstacles, including poor image quality or insufficient grounding capability of verifier, may result in misidentification of hallucinations. 
To enhance the robustness of our method,
we therefore introduce a counterfactual branch.
Specifically, for each entity $e$, we generate a counterfactual description $\tilde{e}$ that is \emph{intended not to be visually co-located} with $e$:
\begin{equation}
    \tilde{e}=\mathcal{C}(e, \mathcal{K}),
\end{equation}
where $\mathcal{C}$ is a counterfactual perturbation function supported by the radiological knowledge $\mathcal{K}$. 
We design $\mathcal{C}$ to satisfy the constraint of \textbf{non-overlapping}: $\tilde{e}$ refers to a mutually exclusive perturbation (e.g., contralateral side, different organ class, or negated attribute) that should not share the same visual evidence region with $e$. For instance, ``left kidney'' versus ``right kidney'', as shown in Fig.~\ref{fig:main}.
This non-overlap guarantee is specifically built into $\mathcal{K}$’s construction: we apply a strict spatial exclusion rule that discards any entity whose bounding box could overlap with others on identical imaging slices.
We then ground the counterfactual as:
\begin{equation}
    (b^{-}_e, s^{-}_e)=\mathcal{G}(I, \text{prompt}(\tilde{e})).
\end{equation}

\subsection{Counterfactual–driven Uncertainty Estimation Algorithm}

We first quantify the overlap between factual and counterfactual grounding:
\begin{equation}
    \mathrm{IoU}_e=\mathrm{IoU}(b^{+}_e, b^{-}_e).
\end{equation}
Intuitively, if the response is correct and the entity $e$ is truly present, the factual grounding should be confident and localized, while the counterfactual should either have low confidence or localize elsewhere. Conversely, if the LVLM model hallucinates $e$ or the grounding verifier $\mathcal{G}$ is confused by the aforementioned obstacles in Sec.~\ref{sec:counterfactual}, both prompts might locate the same spatial region, yielding high overlap that confuses the hallucination decision-making. To address this issue, we novelly combine these signals into an \textbf{IoU-reweighted uncertainty score} $U_e$, which directly measures the likelihood of hallucination:
\begin{equation}
    U_e = 1 - (s^{+}_e - \mathrm{IoU}_e \cdot s^{-}_e).
\end{equation}
This algorithm ensures $U_e$ remains low when factual grounding is confident and spatially distinct from the counterfactual. However, if the LVLM hallucinates $e$ or $\mathcal{G}$ is confused, it may confidently localize to spurious regions, risking missed detection. Here, the counterfactual branch acts as a safeguard: significant overlap penalizes $s^{+}_e$ via the $\mathrm{IoU}_e \cdot s^{-}_e$ term, thereby increasing $U_e$ to flag the hallucination.
We then classify $e$ as hallucinated if $U_e > \tau$, where $\tau$ is a tunable threshold.

\section{Experiments}
\subsection{Experimental Setup}

\textbf{Datasets.}
We construct both training and testing datasets sourced from IMIS-Bench~\cite{cheng2025interactive}, a collection of 80 medical image segmentation datasets covering diverse anatomical regions, where this broad coverage significantly ensures  robust generalization. For \textbf{verifier training}, we extract bounding boxes from segmentation masks and form grounding queries in Qwen3-VL format, resulting in 33,266 pairs.
For \textbf{hallucination detection testing}, we collect 1,904 image-report pairs (958 supporting, 946 hallucinated) generated by four LVLMs: MedGemma-27B~\cite{sellergren2025medgemma}, GPT-5.1~\cite{openai2025gpt5}, Gemini-3-Flash~\cite{team2023gemini}, and Grok-4-Fast~\cite{grok4}, where the hallucinations are induced via mutually exclusive counterfactual labels. This test set spans 1,129 CT and 775 MRI scans.

\textbf{Baselines and Metrics.} As external verifier-based detection is still in its nascent stage, competing methods are limited. Many existing approaches require specialized training on curated datasets or rely on closed-source parameters, rendering direct reproduction infeasible. Consequently, we compare our approach with leading hallucination detection methods, including a GPT-5.1-based prompting method, UniHD~\cite{chen2024unified} and Faithscore~\cite{jing2024faithscore}, where the former two methods are pure API-based and Faithscore relies on trained external expert.
We report the hallucination detection rate (HDR/recall) and precision in Tab.~\ref{tab:main}.

\textbf{Implementation Details.}
We fine-tune a Qwen3-VL-2B model to act as the verifier with LoRA. Specifically, the model is optimized with the AdamW optimizer, using a learning rate of 0.0001 and a weight decay of 0.0001. A CosineLR scheduler is adopted, with the warmup phase accounting for 3\% of the total training steps. The per-device batch size is 16, and gradients are accumulated over 8 steps. Training is conducted for 5 epochs on 2$\times$ NVIDIA A100  GPUs.

\subsection{Main Experimental Results}
To evaluate hallucination detection performance and plug-and-play reusability, we benchmark our detector across multiple LVLM backbones (spanning both general and medical domains) and diverse imaging modalities (CT and MRI), as detailed in Tab.~\ref{tab:main}.
Our method demonstrates robust detection capabilities and consistently outperforms recent baselines across the evaluated backbones and modalities. Notably, on MRI data, our approach achieves superior results on both detection rate and precision. While UniHD reports a higher detection rate on the CT modality, it exhibits the lowest precision among all compared methods, indicating a high rate of false positives (i.e., misidentifying non-hallucinated samples). In contrast, our method maintains the highest precision across all modalities, providing a more reliable balance between sensitivity and accuracy. 
Therefore, the consistent performance across multiple LVLMs validates the plug-and-play capability of our method without requiring access to internal states, which significantly enhances the applicability in real-world clinical settings. Moreover, our method is more lightweight than FaithScore (2B vs. 13B), demonstrating superior computational efficiency.

\begin{table}[t]
\caption{Main experimental results: hallucination detection rate (HDR) and precision (Prec.) across LVLM backbones and imaging modalities (\%).}
\label{tab:main}
\centering
\begin{tabularx}{\textwidth}{l|c|*{8}{>{\centering\arraybackslash}X}}
\hline
\multirow{2}{*}{\textbf{Methods}} & \multirow{2}{*}{\textbf{Param.}}  & \multicolumn{2}{c}{\textbf{MedGemma}} & \multicolumn{2}{c}{\textbf{GPT-5.1}} & \multicolumn{2}{c}{\textbf{Gemini-3}} & \multicolumn{2}{c}{\textbf{Grok-4}} \\
\cline{3-10}
 & & \textbf{HDR} & \textbf{Prec.} & \textbf{HDR} & \textbf{Prec.} & \textbf{HDR} & \textbf{Prec.} & \textbf{HDR} & \textbf{Prec.} \\
\hline
\multicolumn{9}{c}{\textbf{CT}} \\
\hline
GPT-5.1-based  & - & 67.84 & 61.57 & 71.07 & 63.43 & 71.13 & 64.71 & 56.74 & 53.49\\
UniHD~\cite{chen2024unified}   & - & 94.33 & 56.84 & 82.01 & 58.16 & 78.87 & 58.95 & 87.86 & 52.56 \\
Faithscore~\cite{jing2024faithscore}   & 13B   & 58.16 & 61.19 & 48.92 & 64.76 & 52.11 & 64.35 & 55.00 & 65.81 \\
Ours      &  2B    & 63.12 & 64.03 & 61.15 & 68.00 & 62.68 & 60.14 & 60.00 & 59.57 \\
\hline
\multicolumn{9}{c}{\textbf{MRI}} \\
\hline
GPT-5.1-based  & -  & 65.64 & 59.33 & 80.20 & 75.00 & 77.49 & 69.77 & 68.75 & 62.00 \\
UniHD~\cite{chen2024unified}    & -  & 91.58 & 48.33 & 65.66 & 45.46 & 79.79 & 45.18 & 86.46 & 47.43 \\
Faithscore~\cite{jing2024faithscore} & 13B  & 46.32 & 67.69 & 44.44 & 65.67 & 34.04 & 66.67 & 41.67 & 63.49 \\
Ours       &  2B & 95.79 & 91.92 & 91.93 & 92.00 & 96.81 & 90.10 & 95.83 & 93.88 \\
\hline
\end{tabularx}
\end{table}

\subsection{Ablation Studies}
\begin{figure}[t]
  \begin{center}
    \centerline{\includegraphics[width=\columnwidth]{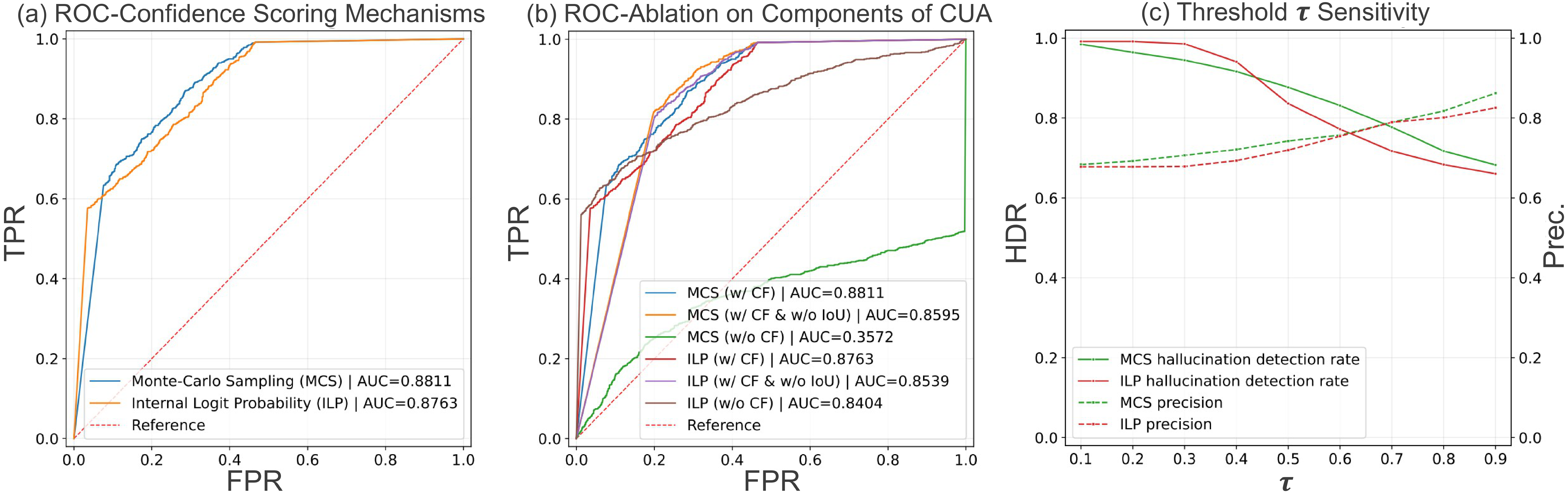}}
    \caption{
    \textbf{Ablation studies on hallucination detection.} 
    (a) Performance comparison of two confidence scoring mechanisms: Monte-Carlo Sampling (MCS) and Internal Logit Probability (ILP). 
    (b) Ablation analysis of the components in Counterfactual-driven Uncertainty Estimation Algorithm (CUA), highlighting the impact of the counterfactual branch (w/CF vs. w/o CF) and IoU-based reweighting (w/CF vs. w/CF \& w/o IoU). 
    (c) Sensitivity analysis of the threshold $\tau$ with respect to HDR and Prec.
    }
    \label{fig:ablation}
  \end{center}
\end{figure}

\textbf{Effect of Different Confidence Scoring Mechanisms.}
We compare the proposed two confidence scoring mechanisms to evaluate their effectiveness in hallucination detection: \textbf{MCS} and \textbf{ILP}. As shown in Fig.~\ref{fig:ablation} (a), both methods achieve high performance, with AUC values of 0.8811 and 0.8763. The MCS method slightly outperforms the ILP, suggesting that evaluating the semantic consistency of multiple outputs is more robust for detecting hallucinations than relying solely on the model's internal logit probability based confidence scores.

\textbf{Effect of Counterfactual--driven Uncertainty Algorithm (CUA).}
To investigate the contribution of each component within our proposed CUA framework, we conduct an ablation study as illustrated in Fig.~\ref{fig:ablation} (b). The results demonstrate that the inclusion of counterfactual (CF) branch is critical for both scoring mechanisms.
Specifically, removing the CF component leads to a severe performance degradation and a notable score-direction inversion for the MCS: the raw AUC drops to 0.3572, which falls below the random guess level. This below-random performance indicates an anti-correlation between the predicted scores and the ground-truth labels. However, when applying a sign flip to correct the prediction direction, the MCS-only baseline achieves a reasonable standalone AUC of 0.6428. This mathematically confirms that while the MCS inherently possesses discriminative capability, the CF branch is indispensable for rectifying the prediction direction and substantially boosting the overall AUC from 0.6428 to 0.8811.
Meanwhile, for the ILP, removing the CF component results in a direct performance decrease from 0.8763 to 0.8404. 
This suggests that CF samples effectively expose the model's underlying uncertainty by creating challenging decision boundaries. Furthermore, we evaluate the role of IoU-based method that reweights the confidence of CF by computing the overlapping of two branches. Removing this filtering step (w/o IoU) results in a moderate decrease in AUC (to 0.8595 for MCS and 0.8539 for ILP), which confirms that filtering out spatially irrelevant samples helps refine the reliability of the uncertainty scores.

\textbf{Sensitiveness to Threshold $\tau$.}
We further analyze how the selection of the threshold $\tau$ impacts the hallucination detection performance. As shown in Fig.~\ref{fig:ablation} (c), as $\tau$ increases from 0.1 to 0.9, the hallucination detection rate for both MCS and ILP exhibits a steady decline. Conversely, the precision of both methods shows a consistent upward trend, reaching its peak at higher $\tau$ values. This reflects a typical trade-off in anomaly detection: a lower threshold allows for the identification of more potential hallucinations, while a higher threshold ensures that the identified instances are more likely to be genuine hallucinations. The intersection of the detection rate and precision curves around $\tau=0.6$ suggests an optimal balance point for general applications.

\subsection{Qualitative Analysis}
\begin{figure}[t]
  \begin{center}
    \centerline{\includegraphics[width=.9\columnwidth]{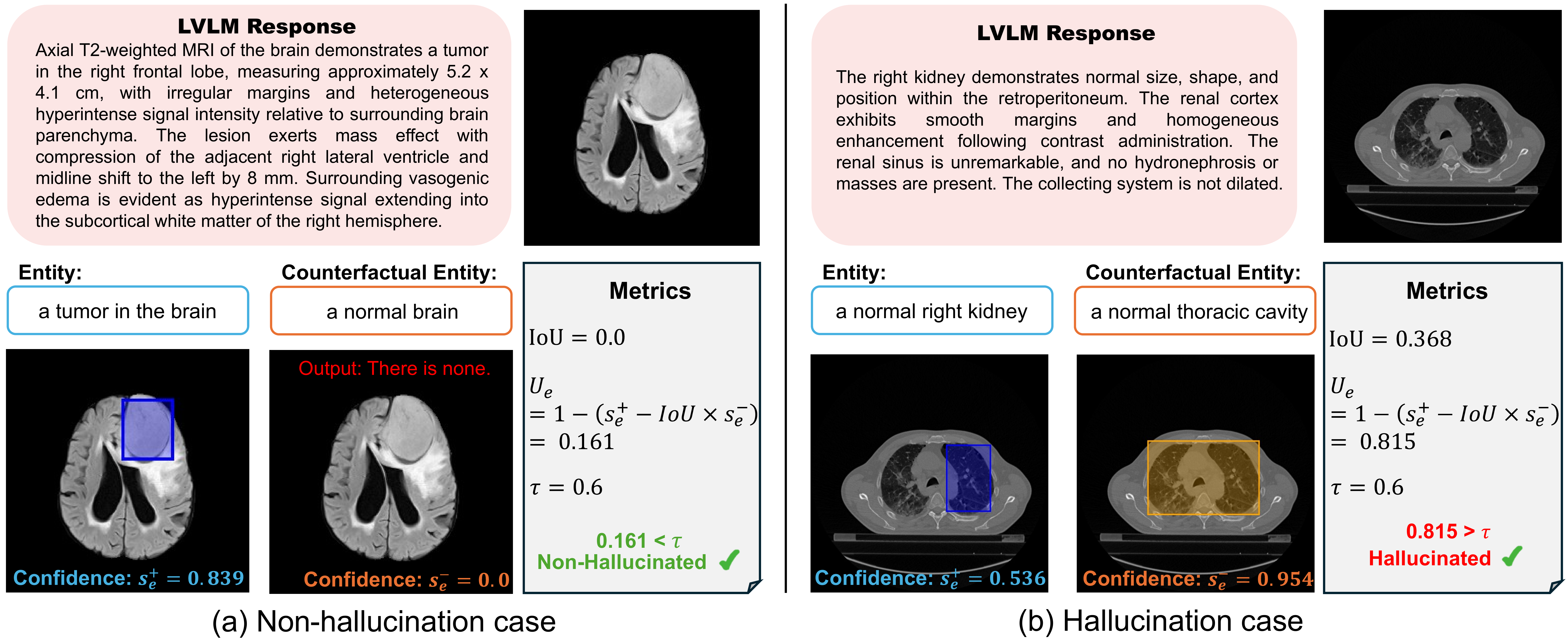}}
    \caption{
    Visualization of hallucination detection. (a) Non-hallucination case where factual branch detects grounded visual evidence and makes correct decision. (b) Hallucination case missed by factual branch but successfully detected by our algorithm.
    }
    \label{fig:vis}
  \end{center}
\end{figure}

Fig.~\ref{fig:vis} shows two hallucination detection cases. Specifically, for the non-hallucinated response ``a tumor in the brain'' in Fig.~\ref{fig:vis}(a), our method performs factual and counterfactual grounding. The verifier accurately localizes the tumor (blue box) with high confidence ($0.839$), while the counterfactual query ``a normal brain'' yields no detection, consistent with pathological fact. This contrast produces a low uncertainty score ($0.161$), confirming factual accuracy. In Fig.~\ref{fig:vis}(b), for ``a normal right kidney'', the verifier fails confident localization ($0.536$), whereas the counterfactual ``thoracic cavity'' yields high confidence ($0.954$). This discrepancy gives $U_e=0.815$, indicating hallucination. Beyond binary detection, our method provides interpretable visual evidence, justifying detection and facilitating targeted rectification for precise response generation.

\section{Conclusion}
We propose a plug-and-play, vision-traceable clinical hallucination detection method for LVLMs via visual grounding experts. To enhance robustness, we introduce a counterfactual entity perturbation branch for uncertainty estimation. This enables robust uncertainty estimation, effectively distinguishing hallucinations by suppressing spurious visual alignments from a single factual branch. Experiments across medical imaging modalities and LVLM backbones demonstrate superior detection performance and provide clinicians with interpretable, entity-level visual localization evidence. Future work will integrate this into hallucination mitigation, extend to complex spatiotemporal relationships in medical imaging, and overcome static, spatially exclusive priors by developing dynamic knowledge bases for visually entangled concepts. Ultimately, this research explores ways for reliable, safety-critical multimodal AI in the clinical domain.

\begin{credits}
\subsubsection{\ackname} This study was funded
by the New Generation Artificial Intelligence-National Science and Technology Major Project (2025ZD0123502).

\subsubsection{\discintname}
The authors have no competing interests to declare that
are relevant to the content of this article.
\end{credits}

\bibliographystyle{splncs04}
\bibliography{Paper-4284}

\end{document}